\documentclass[letterpaper]{article} 
\usepackage[]{aaai25}  
\usepackage{times}  
\usepackage{helvet}  
\usepackage{courier}  
\usepackage[hyphens]{url}  
\usepackage{graphicx} 
\usepackage{natbib}  
\usepackage{caption} 
\usepackage{adjustbox}
\usepackage{algorithm}
\usepackage{algorithmic}
\usepackage{float}

\usepackage{booktabs}       
\usepackage{multirow}

\usepackage{amsmath}
\usepackage{amssymb}
\usepackage{tikz,tikz-cd}
\usetikzlibrary{math,bayesnet,decorations.pathreplacing,shapes.geometric,shapes,trees}
\usepackage{xcolor}
\usepackage{subcaption}
\usepackage{xspace}
\usepackage{placeins}
\usepackage{lipsum}
\usepackage{makecell}
\usepackage{amsthm}
\usepackage{graphicx}
\usepackage{subcaption}

\usepackage{newfloat}
\usepackage{placeins}
\usepackage{listings}
\DeclareCaptionStyle{ruled}{labelfont=normalfont,labelsep=colon,strut=off} 
\floatstyle{ruled}
\newfloat{listing}{tb}{lst}{}
\floatname{listing}{Listing}

\DeclareMathOperator*{\argmin}{arg\,min}
\theoremstyle{definition}
\newtheorem{definition}{Definition}[section]
\title{Beyond Manual Planning: Seating Allocation for Large Organizations}

\author{
  Anton Ipsen\textsuperscript{\rm 1},
  Michael Cashmore\textsuperscript{\rm 1},
  Kirsty Fielding\textsuperscript{\rm 1},\\
  Nicolas Marchesotti\textsuperscript{\rm 1},
  Parisa Zehtabi \textsuperscript{\rm 1},
  Daniele Magazzeni\textsuperscript{\rm 1},
  Manuela Veloso\textsuperscript{\rm 1}
}
\affiliations{
  \textsuperscript{\rm 1}AI Research, J.P. Morgan Chase\\
  \vspace{.3cm}
}

\begin{document}

\maketitle

\section{Abstract}

We introduce the Hierarchical Seating Allocation Problem (HSAP) which addresses the optimal assignment of hierarchically structured organizational teams to physical seating arrangements on a floor plan.
This problem is driven by the necessity for large organizations with large hierarchies to ensure that teams with close hierarchical relationships are seated in proximity to one another, such as ensuring a research group occupies a contiguous area.
Currently, this problem is managed manually leading to infrequent and suboptimal replanning efforts.

To alleviate this manual process, we propose an end-to-end framework to solve the HSAP. A scalable approach to calculate the distance between any pair of seats using a probabilistic road map (PRM) and rapidly-exploring random trees (RRT) which is combined with heuristic search and dynamic programming approach to solve the HSAP using integer programming.
We demonstrate our approach under different sized instances by evaluating the PRM framework and subsequent allocations both quantitatively and qualitatively.

\section{Introduction}
In today's dynamic work environment, commercial office real estate has become one of the most significant fixed costs for companies and organizations. The recent push by many government departments and organizations for a full return to the office has intensified the need to optimize this valuable resource. Unlike other assets, office real estate is not easily expandable due to the prohibitive costs of renovations and lengthy lease agreements. Therefore, the continuous re-optimization of office space usage is essential to maximize efficiency and adapt to changing needs. This ongoing evolution in office space requirements has paved the way for the application of automated planning methods in managing office real estate.

A key challenge in this domain is the Seating Allocation (\textsc{SA}) problem, which involves assigning teams to office space on a floor plan. The \textsc{SA} problem has been previously explored in various contexts, such as allocating physical seats to different parties in government legislative bodies~\cite{Hales2019, Serafini2012}, and in response to social-distancing measures following the COVID-19 pandemic~\cite{barry2021optimal, Stoll2022}.  The objective of the \textsc{SA} problem is to determine an allocation that ensures team members are seated in proximity to one another, where `\textit{proximity}' is defined in terms of walkable space.


The Hierarchical Seating Allocation Problem (\textsc{HSAP}) extends the \textsc{SA} problem by aiming to allocate teams within an organizational hierarchy to a floor plan. This extension is motivated by the need to accommodate teams that belong to multiple departments, thereby linking related teams together. Unlike the \textsc{SA} problem, the objective of the \textsc{HSAP} is to assign teams to office space in such a way that teams closer in the organizational hierarchy are seated closer together within the floor plan. The usefulness between \textsc{HSA} and \textsc{SA} is significant to large organizations as the human-evaluated solution quality to \textsc{SA} would be low due to failure to accommodate linked departments.


The contributions of this paper are as follows: First, we address the critical need for accurate distance estimation between seats by applying a scalable approach that utilizes probabilistic roadmaps and rapidly-exploring random trees to achieve more precise estimations. Secondly, we introduce the Hierarchical Seating Allocation (HSA) problem and demonstrate that it can be effectively addressed by decomposing it into a series of Seating Allocation (SA) sub-problems, which are solved using dynamic programming techniques. Lastly, we extend the existing literature on the SA problem by proposing a local search heuristic aimed at improving the quality of existing solutions. 

The paper is organized as follows.
In Section~\ref{sec:method} we describe how to solve \text{HSA} by solving sub-problems of \textsc{SA}, calculating distances (Section~\ref{sec:method:distance}) and allocating seats (Section~\ref{sec:method:seats}). In Sections~\ref{sec:eval:quant} and \ref{sec:eval:qual} we present our quantitative and qualitative evaluations, and conclude in Section~\ref{sec:conclusion}.

\subsection{Related Problems}
%
A related issue to the SA problem is the Office Space Allocation (OSA) problem. The \textsc{OSA} is the problem of spatially allocating people, entities or teams to a building or floor such that the available space is optimized \cite{ulker2013office, awadallah2012office}.
%
The key distinction between the \textsc{SA} and \textsc{OSA} problems lies in their focus: the \textsc{OSA} problem is concerned with allocating teams to a set of floors based on a set of constraints, while the \textsc{SA} problem, given a valid allocation, focuses on assigning teams to individual seats.

Another closely related problem to SA is the capacitated p-median problem (CPMP). The \textsc{CPMP} involves selecting p-medians (locations) from a finite set of medians to optimally serve a set of demand sites, such that each median does not exceed its associated capacity ~\cite{Serafini2012,Ceselli2005, Boccia2007}. There have been many proposed exact and heuristic approaches to solve the \textsc{CPMP}, including Lagrangian Relaxations~\cite{Lorena2003}, column generation~\cite{Lorena2004}, local search heuristics~\cite{Lorena2003}, and math-heuristics for solving large scale problems~\cite{Reese2006,Stefanello2014,Gngi2021}.
The SA problem can be viewed as a special case of the \textsc{CPMP} where each median has an demand instead of a capacity.

Using this framework the \textsc{SA} problem can be solved by selecting a `\textit{central seat}' (median) of each team, and assigning the required number of seats as close as possible to the assigned `\textit{central seat}' such that each team's desks demands are satisfied. Given that the \textsc{CPMP} is proven to be NP-Hard, utilizing its solution frameworks to solve the \textsc{SA} problem is also NP-hard \cite{Mu2019}. A direct optimization approach could be employed to minimize the total seat distance between each seat, but this results in a mixed-integer quadratic problem (MIQP), which is not scalable for large instances due to the number of decision variables in the objective. A detailed formulation is provided in Appendix 1.


\section{Framework} \label{sec:method}

We will discuss in detail our proposed end-to-end framework, which is divided into two key tasks: 1) calculating the pairwise seat distances, and 2) allocating teams to specific seats within the floor plan. The importance of these two tasks is evident in the objective of the \textsc{CPMP}, where the goal is to minimize the distance between seats and their designated `\textit{central seat}'. Therefore, a pairwise seat distance matrix has to be calculated as it serves as the input for the subsequent allocation process. If the estimated distances between seats are either overestimated or underestimated, it can lead to illogical allocations when evaluated by a human decision-maker. Conversely, if the allocation framework in step 2 does not adequately model human preferences, the solution quality, as reviewed by a human, would similarly be poor.

The optimization of allocating teams to seats is formulated as a combination of mathematical heuristics and Mixed Integer Programming problems (Section~\ref{sec:method:seats}). We will model this as a multi-resource demand problem with two distinct desk types: desks and single-occupant offices. Single-occupant offices are important in many organizations, as high-ranking employees are usually allocated these spaces. Going forward, we will refer to single-occupant offices simply as offices.

\subsection{Estimating Pairwise Seat Distances}\label{sec:method:distance}

In this section, we present a method for obtaining high-quality estimates of pairwise seat distances as identified on floor plans. Following the approach of \citet{barry2021optimal}, we employ computer vision techniques and object detection to parse floor plans, determining and mapping the locations of desks and offices in the $x,y$-plane.

A naive approach to estimating walking distances between each pair of seats (including both desks and offices) would be to use Manhattan or Euclidean distance. While these methods are computationally efficient, they fail to account for non-traversable areas within a floor plan. This oversight can lead to severe distance underestimation, potentially resulting in team allocations on opposite sides of walls, thereby significantly diminishing the quality of solutions as evaluated by human experts.

Grid-based pathfinding algorithms such as $A^*$ \cite{A*} and FastMap \cite{cohen2017fastmapalgorithmshortestpath} exist, as do other efficient optimal mesh-based algorithms \cite{hechenberger2022multirayScan,polyanya}. However, the aforementioned approaches require an efficient and sparse representation of the floor plan in order to calculate the pairwise seat distances fast \cite{Wu2021}. For high resolution floor plans when projected into $x,y$-plane, obtaining such efficient representations may require substantial manual editing. Similarly, finding an adequate representation is also made difficult if the quality of the provided floor plans is low. Moreover, for large organizations with extensive office footprints, manually adjusting each representation to compute optimal pairwise seat distances may be infeasible.

To overcome the representation challenges of exact methods and the underestimation issues of distance norm approaches, we propose constructing a \textit{probabilistic roadmap} (\textsc{PRM}) using rapidly-exploring random trees (RRT), augmented with an additional connectivity check that automatically connects seats within a threshold distance from any reachable node of the \textsc{PRM}. In this context, a node within the \textsc{PRM} can represent either a point used for space exploration, a desk, or a single-occupant office. This yields a fully-connected, undirected graph that effectively approximates the walking distance between every pair of seats~\cite{kav96_prm,lav98_rrt}. The process for generating the \textsc{PRM} that connects a set of seat locations $S$ with at most $K$ additional nodes is described in Algorithm~\ref{alg:prm}.

\begin{algorithm}
    \caption{Generate PRM}
    \label{alg:prm}
    \begin{algorithmic}[1]
    \STATE \textbf{Input:} $K,S,\delta_c,\delta_s$
    \STATE \textbf{Output:} PRM $\langle N,E \rangle$
    \STATE $x_{init}\leftarrow$ random collision-free position
    \STATE $N,E \leftarrow \{x_{init}\}, \emptyset$
    \FORALL{$s\in S$}
        \STATE $connected \leftarrow$ connect\_seat$(s,N,E,\delta_s)$
        \WHILE{$\neg connected$ and $|N| < K+|S|$}
            \STATE $x_{rand}\leftarrow$ random collision-free position
            \STATE $n_{nearest}\leftarrow$nearest\_neighbour$(N,x_{rand})$
            \STATE $x_{new}\leftarrow$ cast$(n_{nearest},x_{rand},\delta_c)$
            \IF{$x_{new}\neq\bot$}
                \STATE $E\leftarrow E\cup$ new\_edges$(x_{new}, N)$
                \STATE $N\leftarrow N\cup\{x_{new}\}$
                \STATE $connected \leftarrow$ connect\_seat$(s,\{x_{new}\},E,\delta_s)$
            \ENDIF
        \ENDWHILE
    \ENDFOR
    \RETURN $\langle N,E \rangle$
  \end{algorithmic}  
\end{algorithm}

\begin{figure*}[t]
    \centering
    \includegraphics[width=0.95\textwidth]{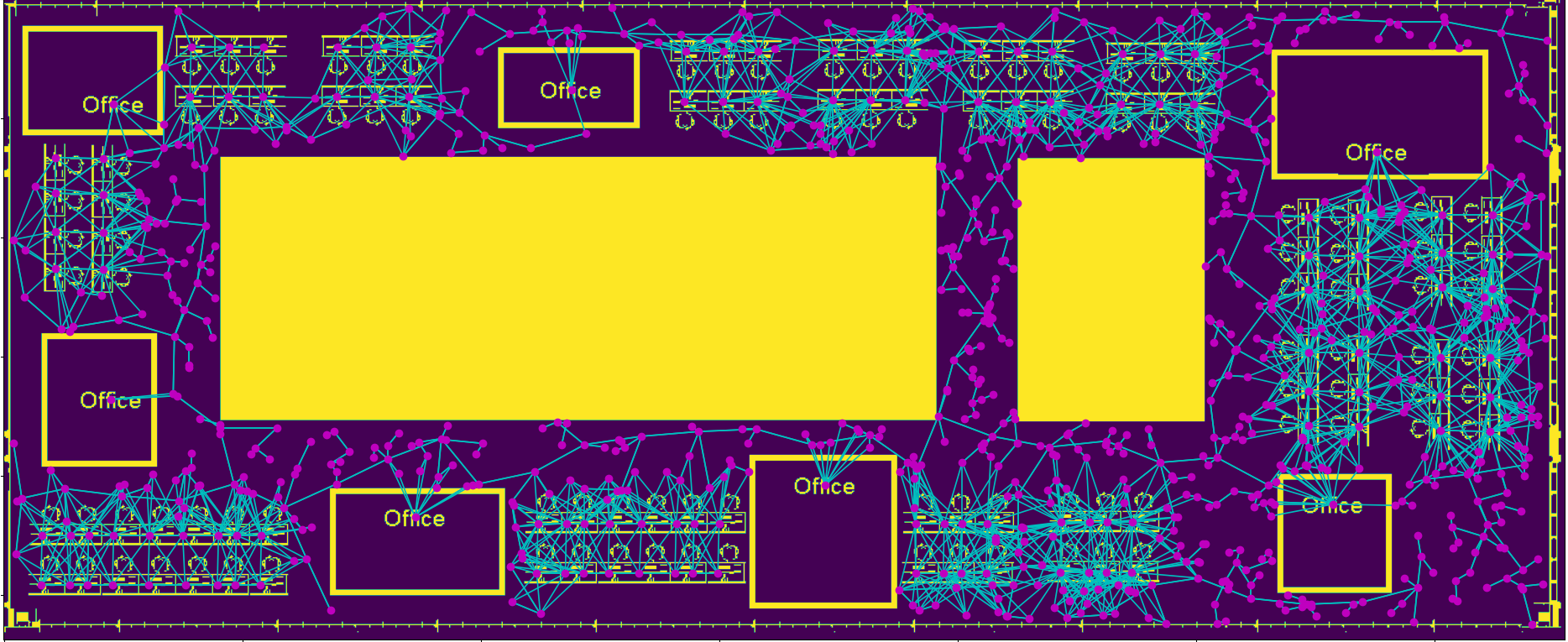}
    \caption{Output of the \textsc{PRM} algorithm when applied to a synthetic floor plan. The hyper-parameters for $\delta_c$ and $\delta_s$ were manually tuned until the algorithm converged with a node limit of $K=2000$.
    }
    \label{fig:RRT}
\end{figure*}
The procedure is initialized with a set of seat positions $S\in\mathbb{R}^2$ representing both desks and offices; a maximum number of additional nodes $K$ and distance thresholds $\delta_c,\delta_s\in\mathbb{R}$.
The algorithm outputs a connected graph of nodes $N$, corresponding to seats and intermediate positions, and edges $E$, representing collision-free connections between nodes.
The \textsc{PRM} is initialized as a single node randomly sampled from any collision-free position on the floor plan (line 3).
 
If all seats are connected, or the maximum number of nodes is reached, the algorithm returns the resulting \textsc{PRM}, as illustrated in Figure~\ref{fig:RRT}. If the algorithm fails to connect all the seats, $K$ can be increased, and the process repeated.
Once all seats are successfully connected, the \textsc{PRM} forms a connected graph. This allows for the efficient calculation of the shortest pairwise distances between seats using the Floyd-Warshall algorithm, which operates with a time complexity of $O(n^3)$.

The theoretical lower bound for the proposed \textsc{PRM} method is the Euclidean distance. Since seats are connected within a $\delta_s$ radius, and assuming $\delta_s = \infty$ the method is lower bounded by the Euclidean distance, as every node would be connected by the shortest possible path.

In the next section we will present several approaches to solve the \textsc{SA} problem and the solution to the \textsc{HSA} problem.

\subsection{Constrained Seat Allocation}\label{sec:method:seats}
The basic \textsc{SA} problem is defined as follows.
\begin{definition}[Seat Allocation (SA)] A \textbf{\textsc{SA} problem} is a tuple $\langle S, T, D \rangle$, in which $S := S_d \cup S_o$ is the set of available seats (desks and offices). $T$ is the set of teams. Each team $t\in T$ has an associated desk requirement $d_t\in \mathbb{N}$ and office requirement $o_t\in \mathbb{N}$. Lastly, $D:s\times s\rightarrow \mathbb{R}_{\geq 0}$ is the set of pairwise distances between seats.
\end{definition}
A solution to a \textsc{SA} problem is an assignment $X:S\rightarrow T$, such that for each team $t$ the number of desks and offices assigned meets its requirements. We measure the quality of a solution by measuring the total distance between each seat assigned to the team and the central seat of that team. That is:
$$
\sum_{t\in T} \sum_{s : X(s)=t} D(s,s'_t)
$$
where $s'_t$ is the central seat for team $t$, i.e. the seat that minimizes:
$$
\argmin_{s'\in S : X(s')=t} \sum_{s : X(s)=t} D(s,s'_t)
$$
There are several other ways of measuring the quality of solution such as minimizing the the maximum distance to any given central seat.

\subsubsection{Integer Program Seat Allocation (IPSA)}
To solve the \textsc{SA} we present an Integer Program (IP) that is adapted from Hales and Garcia ~\shortcite{Hales2019} to include both desks and offices.
The idea of the IP is to find a central seat (centroid) for each team and allocating required number of desks and offices to the corresponding central seat to satisfy the associated team's demand.
We define the binary variable $x_{ik}$ as $1$ if seat $i$ is allocated to central seat $k$, and the binary variable $y_{kt}$ as $1$ if seat $k$ is the central seat for team $t$. Moreover, we denote $D(i,k)$ as the distance between seats $i$ and $k$ as calculated by the \textsc{PRM} approach described in the section~\ref{sec:method:distance}. We denote the following method as Integer Program Seat Allocation (IPSA).
%
\begin{align}
    \min \sum_{i \in S, k\in S} D(i,k) x_{ik} \label{eq:IPObjective} \\
    \text{s.t.} \sum_{k\in S} x_{ik}\le 1 \quad \forall i \in S \label{eq:allseatsCentralSeat} \\
    \sum_{t \in T} y_{kt} \leq 1 \quad \forall k \in S \label{eq:SeatOneCentral} \\
    \sum_{k \in S} y_{kt} = 1 \quad \forall t \in T \label{eq:oneCentralSeatTeam} \\
    \sum_{i\in S_d} x_{ik}=\sum_{t \in T} d_t y_{kt} \quad \forall k \in S \label{eq:TeamDeskCapacity} \\
    \sum_{i\in S_o} x_{ik}=\sum_{t \in T} o_t y_{kt} \quad \forall k \in S \label{eq:TeamOfficeCapacity}
\end{align}

The objective function in Eq.~\eqref{eq:IPObjective} minimizes the distances between the seats allocated to their central seats. The constraints can be explained as follows:
\begin{itemize}
    \item Eq.~\eqref{eq:allseatsCentralSeat} stipulates that all seats are assigned to at most one central seat. The equality sign is less-or-equal due to scenarios where the total seat supply is higher total seat demand.
    \item Eq.~\eqref{eq:SeatOneCentral} describes that seat $k$ can only be the central seat of at most one team and Eq.~\eqref{eq:oneCentralSeatTeam} describes that each team should have exactly one central seat.
    \item Eq.~\eqref{eq:TeamDeskCapacity} and Eq.~\eqref{eq:TeamOfficeCapacity} stipulate that if seat $k$ is central for team $t$, then the required number of desks and offices for team $t$ must be assigned to central seat $k$.
\end{itemize}

The \textsc{IPSA} approach is computational expensive in practice due to the rapidly increasing number of variables in $x_{ik}$ ($|S|^2$), which makes large-scale problems intractable. This motivates the need for heuristics to reduce the computationally complexity.

\subsubsection{Iterative Clustering Algorithm (ICA)}\label{seq:ICA}

To bridge the gap between optimality and scalability, we present a heuristic approach to \textsc{SA} that decomposes the problem into two phases: (1) selecting central seats, (2) allocating seats to teams. We refer to these phases as \textit{location} and \textit{allocation}, respectively~\cite{Hales2019}.
The heuristic approach closely resembles an iterative k-means clustering algorithm. In each iteration, the centroid of each cluster is updated (\textit{location}), followed by the allocation of seats to their nearest centroid (\textit{allocation}). The overall approach is described in Algorithm~\ref{alg:LAH}.

Before any iteration begins, an initial central seat is assigned to each team (line 3). This is done by one of two methods: randomization or k-means++~\cite{kmeans++2007}. We refer to the first approach as \textsc{ICA}, and the second as \textsc{ICA++}.
To solve the \textit{allocation} problem, we use an integer programming approach (line 7). We begin by defining $s_t$ as the central seat allocated to team $t$. Additionally, we introduce a binary variable $x_{it}$, which is $1$ if seat $i$ is allocated to team $t$. This leads us to the following optimization problem:

\begin{align}
    \min \sum_{i \in S, t \in T} D(i,s_t) x_{it} \label{eq:ICAObjective} \\
    \text{s.t.} \sum_{t\in T} x_{it}\leq 1 \quad \forall i \in S \label{eq:ICAallseatsCentralSeat} \\
    \sum_{i\in S_d} x_{it} = d_t  \quad \forall t \in T \label{eq:ICATeamDeskCapacity} \\
    \sum_{i\in S_o} x_{it} = o_t \quad \forall t \in T \label{eq:ICATeamOfficeCapacity}
\end{align}

The constraints and objective function are similar to those defined for the \textsc{IPSA}.
\begin{itemize}
    \item The objective function~\eqref{eq:ICAObjective} minimizes the sum of distances between each seat and its central seat.
    \item Eq.~\eqref{eq:ICAallseatsCentralSeat} stipulates that all seats are allocated to at most one team.
    \item Eq.~\eqref{eq:ICATeamDeskCapacity} and Eq.~\eqref{eq:ICATeamOfficeCapacity} stipulate that each team must be allocated the required desks and offices.
\end{itemize}

The \textit{location} phase updates the location of the central seats to minimize~\eqref{eq:ICAObjective}. This is done by assigning a new central seat to each team  that minimizes:
$$\Bar{S_t}= \argmin_{s \in S} \sum_{i\in S}D(i,s)x_{it}$$
That is, the new central seat is the seat already allocated to that team which has the smallest total distance to all other seats allocated to that team (line 8).

This process repeats until the central seats remain unchanged (lines 9-10) or the maximum number of repetitions is reached, at which point the iteration stops and the current allocation is returned. The ICA is easier to solve than the IPSA due to fewer decision variables and number of variables in $x_{it}$ is $(|T| \times |S|)$ as $|T|\leq |S|$ for all feasible problems.

\begin{algorithm}
    \caption{Location-Allocation Heuristic }
    \begin{algorithmic}[1]
        \STATE \textbf{Input:} Seats $S$, teams $T$,distance matrix $D$,init,$max\_iterations$
        \STATE \textbf{Output:} allocation $X$
        \STATE $ct \leftarrow 1$
        \STATE $X \leftarrow \emptyset$    
        \STATE $S_t \leftarrow init(S,T)$
        \WHILE{$ct \leq max\_iterations$}
            \STATE $X \leftarrow allocation(S,T,D,S_t)$
            \STATE $\Bar{S_t} \leftarrow location(S,T,D,X)$
            \IF{$\Bar{S_t} = S_t \ \forall t \in T $}
                \RETURN $X$
            \ENDIF
            \STATE $S_t \leftarrow \Bar{S_t}$
            \STATE $ct \leftarrow ct + 1$
        \ENDWHILE
        \RETURN $X$
    \end{algorithmic}
    \label{alg:LAH}
\end{algorithm}

\subsubsection{Greedy Seat Allocation (GSA)}

In this subsection we present a greedy algorithm as a tractable baseline to the \textsc{IPSA} method. \textsc{GSA} algorithm allocates desks and offices to teams based on the capacity of selected central seats in a greedy manner.
The initial set of central seats is allocated using a k-means++ clustering approach~\cite{kmeans++2007}. Then the subsequent \textit{allocation} phase of the greedy approach is an adapted version of a capacity-based greedy method proposed by Gn\"agi and Baumann~\shortcite{Gngi2021}.
The detailed steps of this allocation process are outlined in Algorithm~\ref{alg:greedy}.

The assignment order of each seat is determined using the regret function proposed by Mulvey and Beck~\shortcite{Mulvey1984}.  For each seat $s$, we compute the regret value $r_s = D(s,c')-D(s,c'')$, where $c'$ and $c''$ are the first and second nearest central seat, respectively (lines 5-6).
 A higher regret value suggests a greater potential increase in the objective function if the seat were reassigned to a different central seat.
Next, the \textit{assignment\_order} function calculates the order in which teams are considered for seat $s$. Teams are prioritized based on their proximity to the seat's central seat (line 9), ensuring that the seat is allocated to the nearest possible team.
Finally, each seat $s$ is iteratively evaluated for allocation to team $t$ contingent on whether the team's office and desk requirements have been fulfilled.The allocation sequence is guided by the regret values, and for each seat, team consideration follows the assignment order (lines 10-21).

Once the \textit{allocation} phase is complete the central seats are relocated (Algorithm~\ref{alg:LAH} line 8). The process repeats until we hit the maximum number of iterations, or the location of central seats has stabilized at which point the algorithm outputs a solution, $X$ (Algorithm~\ref{alg:LAH} lines 6-14).

\begin{algorithm}
    \caption{Greedy Allocation}
    \begin{algorithmic}[1]
        \STATE \textbf{Input:} Seats $S$,teams $T$,distance matrix $D$
        \STATE \textbf{Output:} Seating Allocation $X$
        \STATE $S_t \leftarrow \text{initial central seats $s_t \in S$ for each team, $t$}$
        \STATE $X \leftarrow \emptyset$
        \FOR{$\text{seat} \  s \in S$}
            \STATE $r_s \leftarrow D(s,c')-D(s,c'')$
        \ENDFOR
        \FOR{$s\in S$ ordered by decreasing $r_s$}
            \STATE $A_t \leftarrow \text{assignment\_order}(s,S_t,T,D)$
            \FOR{$t \in A_t$}
                \IF{$s\in S_o \wedge o_t > 0$}
                    \STATE $X(s)=t$
                    \STATE$o_t \leftarrow o_t -1$
                    \STATE \textbf{break}
                \ELSIF{$s\in S_d \wedge d_t > 0$}
                    \STATE $X(s)=t$
                    \STATE$d_t \leftarrow d_t -1$
                    \STATE \textbf{break}
                \ENDIF
            \ENDFOR       
        \ENDFOR
        \RETURN $X$
    \end{algorithmic}
    \label{alg:greedy}
\end{algorithm}

\subsubsection{Local Search Heuristic (LS)}

In this subsection, we propose a simple local neighborhood search heuristic (\textsc{LS}) to improve the solutions generated in the previous sections.
The local search heuristic starts from an initial incumbent solution $X^*$. \textsc{LS} attempts to solve the \textsc{IPSA} problem by only considering, for any current central seat in $X^*$, the $s_{n}$-nearest neighbors seats as new candidates for central seats.
Additionally, since $X^*$ is a valid feasible solution, it is used as a warm start for the reduced-dimensionality \textsc{IPSA} problem.
In this way, \textsc{LS} will only improve the current solution, or will terminate with the current solution as the best incumbent. Additionally, the \textsc{LS} method has $ s_{n} \times |T|$ number of candidates central seat which in many instances is less than  $|S| \times |T|$.
The notation of each decision variable follows that of the \textsc{IPSA} method.
We denote $S^*_c$ as the set of seats that are $s_{n}$-nearest neighbors to the central seats selected in $X^*$. Then we can define the local search optimization procedure as:

\begin{align}
    \min \sum_{i \in I, k\in S_c^*} D(i,k) x_{ik} \label{eq:LSIPObjective} \\ 
    \text{s.t.} \sum_{k\in S_c^*} x_{ik}\leq 1 \quad \forall i \in S \label{eq:LSallseatsCentralSeat} \\
    \sum_{i\in S_d} x_{ik}=\sum_{t \in T} d_t y_{kt} \quad \forall k \in S_c^* \label{eq:LSTeamDeskCapacity} \\
    \sum_{i\in S_o} x_{ik}=\sum_{t \in T} o_t y_{kt} \quad \forall k \in S_c^* \label{eq:LSTeamOfficeCapacity} \\
    \sum_{t \in T} y_{kt} \leq 1 \quad \forall k \in S_c^* \label{eq:LSSeatOneCentral} \\
    \sum_{k \in S_c^*} y_{kt} = 1 \quad \forall t \in T \label{eq:LSoneCentralSeatTeam}
\end{align}
The constraints and objective function in the above formulation follows exactly the definition of \textsc{IPSA}.

\subsection{Hierarchical Seat Allocation}\label{sec:HSA}

\begin{figure}[t]
    \centering
    \resizebox{0.6\columnwidth}{!}{%
        \begin{tikzpicture}[%
          grow via three points={one child at (0.5,-0.7) and
          two children at (0.5,-0.7) and (0.5,-1.4)},
          edge from parent path={(\tikzparentnode.south) |- (\tikzchildnode.west)}]
            \tikzset{every node/.style={draw=black,thick,anchor=west}}
            \tikzset{selected/.style={draw=red,fill=red!30}}
            \tikzset{optional/.style={dashed,fill=gray!50}}
            \node {Company A}
            child { node {Sales}
                child { node {EU Sales}}
                child { node {US Sales}}
                child { node {Sales Engineering}}       
            }		
            child [missing] {}
            child [missing] {}
            child [missing] {}
            child { node {Engineering}
                child { node {R\&D Engineering}}
            }
            child [missing] {}
            child { node {Marketing}
                 child { node {Ads}}
                 child { node {Digital}}
            }
            ;
        \end{tikzpicture}%
    }
    \caption{An example organizational hierarchy. Each node corresponds to a team and each child node is a member of its own team and its parent team.}
    \label{fig:hierarchy}
\end{figure}
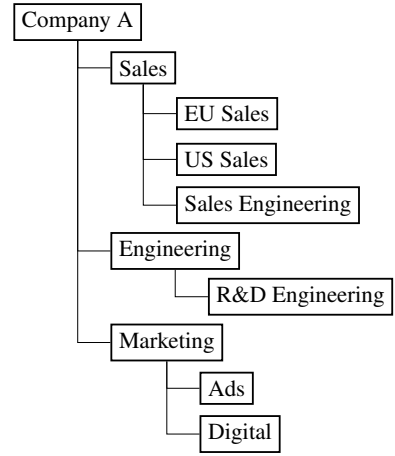

 Consider the organizational hierarchy as depicted in Figure~\ref{fig:hierarchy}. Company A is considered the root team, whereas Sales is a parent team along with Engineering and Marketing. Sales has several child teams that we aim to have the members seated together. A solution must allocate seats to the child teams such that their office and desk requirements are satisfied. Additionally, the solution must minimize the distance between the assigned seats of child teams of each parent team. For instance in Figure~\ref{fig:hierarchy}, it is desired that the distance between EU Sales, US Sales and Sales Engineering must be minimized, and thus their seats must come from the seats allocated to its parent team, Sales. 


The \textsc{HSAP} is mathematically defined as follows.

\begin{definition}[Hierarchical Seat Allocation Problem (HSAP)] An \textbf{HSAP} is a tuple $\langle S, \mathcal{H}, D \rangle$, in which
$S := S_d \cup S_o$ is the set of available seats (desks and offices).
$\mathcal{H}$ is the organizational hierarchy with $H$ levels in which $T_{l}$ denotes the set of teams at level $l$.
Each team $t\in T_l$ has an associated desk requirement $d_t\in \mathbb{N}$, office requirement $o_t\in \mathbb{N}$, and when $l>0$ a parent team $t'\in T_{l-1}$.
Lastly, $D:s\times s\rightarrow \mathbb{R}_{\geq 0}$ is the set of pairwise distances between seats.
\end{definition}

A team with no child teams is referred to as a \textit{leaf team}. Note that the office and desk requirements of non-leaf teams (\textit{branch teams}) are equal to the sum of the requirements of their child teams.

A team without a parent is denoted as a \textit{root team}. Note that it is possible that the hierarchy has several root teams, and could be represented as a set of trees.

A solution to an \textsc{HSAP} is a set of allocations $X:=\{X_0,\ldots X_L\}$. Each allocation $X_l\in X$ is a mapping $X_l: S\rightarrow T_l$ from seats to teams at that level, such that for each team $t\in X_l$ the number of desks and offices assigned to that team meets its requirements. Also, the assignment to a non-root team in level $l+1$ must not violate that of its parent at level $t$. That is, teams at level $l+1$ must be assigned to seats that were previously allocated to their parent at level $l$.

The solution quality for \textsc{HSAP} is measured in the same way as \textsc{SA} (total distance of each seat to its team's central seat). However, for the \textsc{HSAP} we take the sum of this metric at every level in the hierarchy.
Importantly, the branch teams serve as a mechanism of creating structure and synergy in the hierarchical allocations, whereas the leaf teams are the teams that functionally have to be located as close as possible. 

Solving the \textsc{HSAP} problem as a single monolithic instance is intractable for small, medium and large scale instances (i.e. over 50 seats).
Therefore, we propose a decomposition of the hierarchical seating problem into a series of \textsc{SA} subproblems that can be solved independently using the proposed heuristics in Section~\ref{sec:method:seats}.

\begin{figure*}[h]
    \centering
    \begin{subfigure}[t]{0.42\textwidth}
        \centering
        \includegraphics[width=\textwidth]{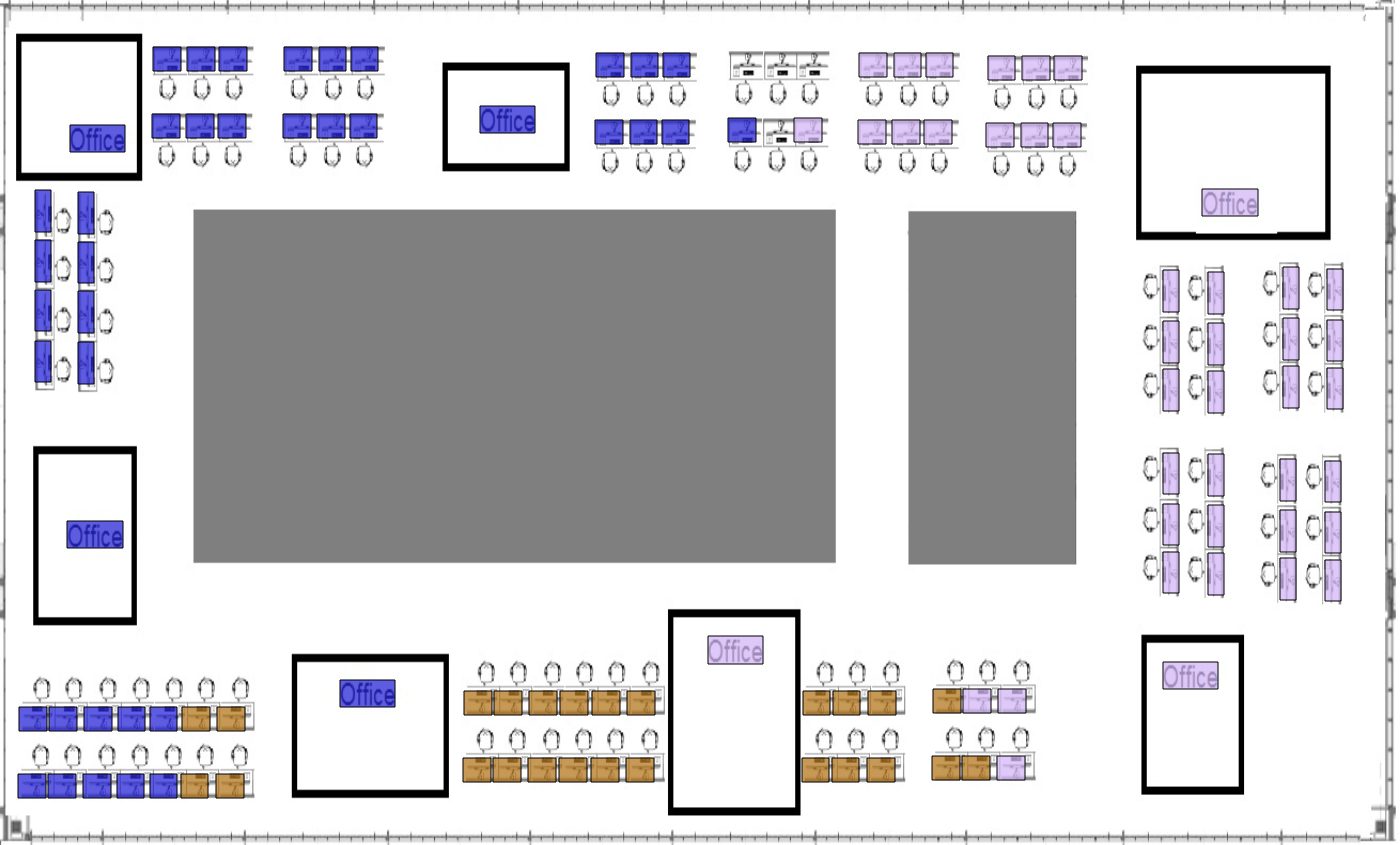}
        \caption{Hierarchy Level 1}
        \label{fig:H1}
    \end{subfigure} 
    \begin{subfigure}[t]{0.42\textwidth}
        \centering
        \includegraphics[width=\textwidth]{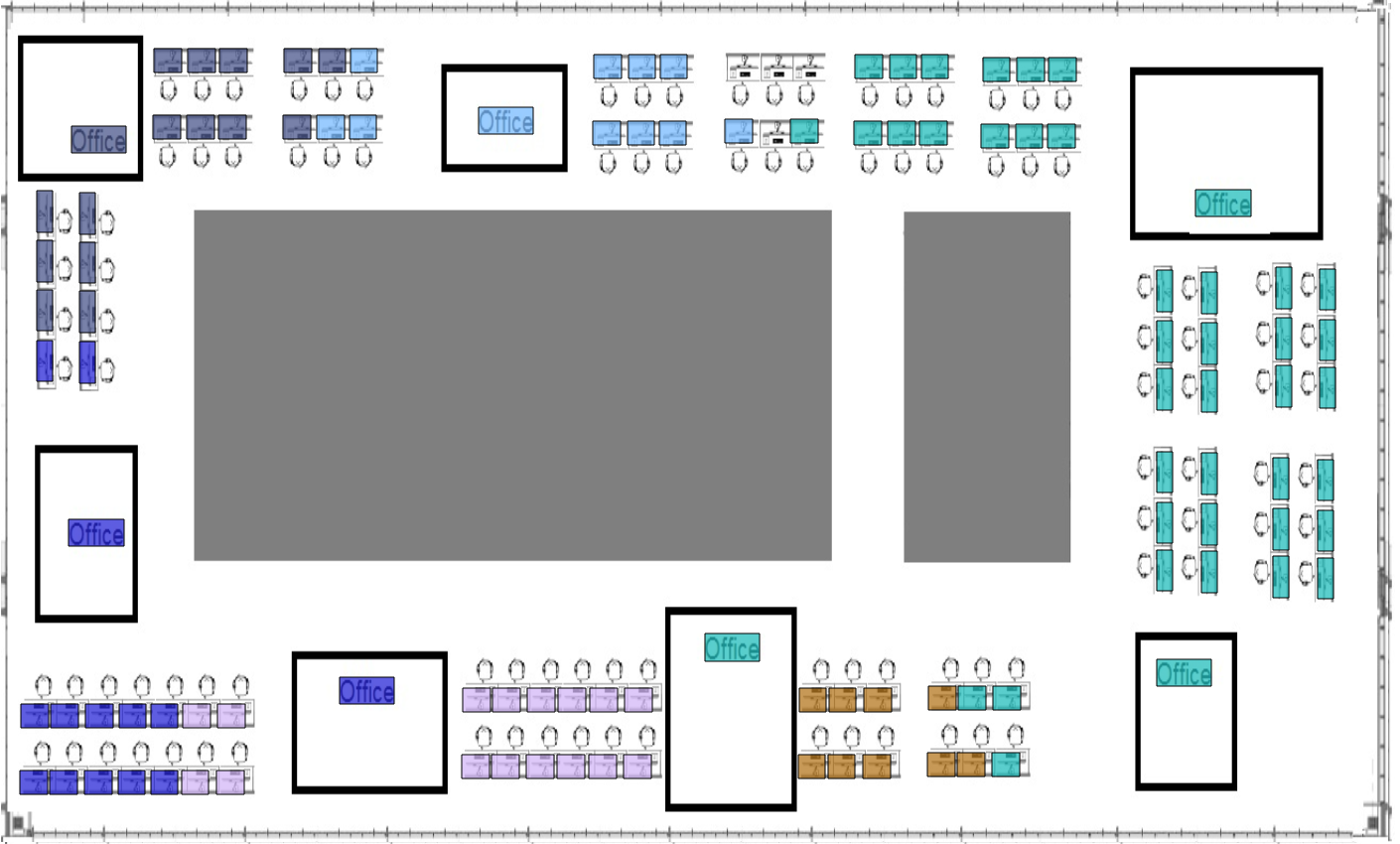}
        \caption{Hierarchy Level 2}
        \label{fig:H2}
    \end{subfigure}
   \begin{subfigure}[t]{0.13\textwidth}
        \centering
        \includegraphics[width=\textwidth]{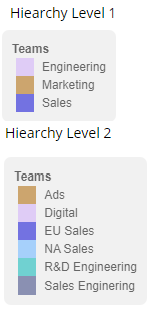}
        \caption{Color legend}
        \label{fig:270_Park_delayed}
    \end{subfigure}
    \caption{Depicts the output of the \textsc{HSA-DF} algorithm with local search heuristics as applied to the organizational hierarchy as described in Figure~\ref{fig:hierarchy} with the floor plan from Figure~\ref{fig:RRT}. In total the population required 103 desks and 6 offices, leaving 3 vacant desks.  }
    \label{fig:output}
\end{figure*}

\subsubsection{Depth First Hierarchical Seat Allocation (DF-HSA)}\label{sec:DF-HSA}
 
To address the challenge of allocating an organizational hierarchy of teams we introduce the \textsc{DF-HSA} algorithm.
As the allocation of each team in level $l$ is constrained by the allocation of seats to teams in level $l-1$, the problem naturally decomposes into a set of \textsc{SA} problems that can be solved sequentially.
Thus, we use a dynamic programming approach to allocate the hierarchy in a top-down, depth-first approach, in which each sub-problem involves solving a single \textsc{SA} instance.

\begin{algorithm}
    \caption{Depth First Hierarchical Seat Allocation (DF-HSA)}
    \label{alg:HSA}
    \begin{algorithmic}[1]
        \STATE \textbf{Input:} Seating allocation algorithm $SA$, seats $S$, teams $T$,distance matrix $D$,previous allocation $X$,iteration $l$
        \STATE $X_l \leftarrow X_l \cup SA(S,T,D)$
        \FOR{team $t\in T$}
            \STATE $S' \leftarrow \{s:X_l(s)=t\}$
            \STATE $T' \leftarrow children(t)$
            \IF{$T' \neq \emptyset$}
                \STATE Call $DF-HSA(SA,S',T',D,X,l+1)$
            \ENDIF
        \ENDFOR
     \RETURN $X_l$
    \end{algorithmic}
\end{algorithm}

The algorithm is initially called with a set of empty allocations ($X_l=\emptyset, \forall X_l\in X$), root teams $T$, and level $l=0$. Seats are allocated by a recursive (depth-first) traversal through the organizational hierarchy (line 3-9).
At each step of this traversal, teams are first allocated seats (line 2). This allocation is performed using any solution to \textsc{SA}, such as those described in the previous section.
Then, each team's children are allocated seats through a recursive call (line 7). The children of a team (line 5) can only be assigned seats that were allocated to their parent (line 4).

The \textsc{DF-HSA} algorithm is not globally optimizing every level at once as each level builds on the previous level's allocation.
\subsubsection{Delayed Office Selection}

When allocating offices in the first iteration of the \textsc{DF-HSA} poor allocation choices can be made in the first levels of the hierarchy due to the lack of knowledge as to which leaf teams those offices are aligned. 
We observe that this leads to allocations in which offices are allocated far from their team, but there were obvious better selections to make if there had been more information about office alignment while allocating higher levels of the organizational hierarchy.

We tackle this issue by the delaying office allocation decision.
We propose a combined top-down-bottom-up approach, in which first desks are allocated using the \textsc{HSA} scheme described above, and then offices are allocated at the lowest level of leaf teams.

From the \textsc{DF-HSA} procedure, we can also extrapolate the central seats for each team, which we can use to allocate office as close to the central seat of the teams as possible.

The office allocation step is a simple form of the ICA approach. We define $x_{it}$ as 1 if office $i$ is allocated to team $t \in T$. We denote $\mathcal{S}_O$, as the set of available offices. Similarly to Section~\ref{seq:ICA} we already know for each team the central seat $s_t$. Lastly, we denote $o_t$ as the office requirements of team $t$.

\begin{align}
    \min & \sum_{i \in S_o, t \in T} D(i, s_t) x_{it} \label{eq: DO Objective} \\
    \text{s.t.} \quad & \sum_{t \in T} x_{it} \leq 1 \quad \forall i \in \mathcal{S}_o \label{eq:DOallseatsCentralSeat} \\
    & \sum_{i \in S_o} x_{it} = o_t \quad \forall t \in T \label{eq:DOTeamOfficeCapacity}
\end{align}

The constraints closely follow the \textsc{ICA} approach, and thus \eqref{eq: DO Objective} defines the total distance from offices to their central seats, \eqref{eq:DOallseatsCentralSeat} defines that every office can only be allocated to one team and \eqref{eq:DOTeamOfficeCapacity} defines that each team must be allocated offices equal to its requirements.

\section{Evaluation}\label{sec:eval:quant}

We assess the performance of the proposed methods from both a quantitative and qualitative perspective. Quantitatively, we report the mean distance from a seat to the corresponding central seat, averaged over all levels.
We measure solution quality in three different floor plan instances to gauge the efficacy for different teams sizes and floor complexity.
The organizational hierarchy for each instance encompasses up to 5 different levels:
\begin{itemize}
    \item Small Instance: Capacity of floor: 172 desks and 26 offices. Team Requirements: 167 desks and 12 offices
    \item Medium Instance: Capacity of floor: 275 desks and 27 offices. Team requirements: 245 desks and 23 offices.
    \item Large Instance: Capacity of floor: 966 desks and 31 offices. Team requirements: 843 desks and 22 offices.
\end{itemize}

Note that the unit of the distance reported is pixels (of the floor plan image) and as a result we do not compare distances between floor plans. 
We ran the solver using Python~3.8 and CPLEX~12.9~\cite{cplex2017v12}. Each instance is solved by considering a total of 5 levels from the organizational hierarchy, with results averaged over 30 runs.

\subsection{Quantitative Evaluation}

We compare results of solving the \textsc{SA} sub-problems with the different \textsc{SA} methods defined in Section~\ref{sec:method:seats}. The setting of the evaluation is that of the human-decision maker. The task to generate seating allocation plans itself is an iterative process. The human decision maker will generate different solutions for a given allocation and then adjust those solutions continuously in response to feedback, changing business constraints, and in the case where the building is being constructed – changing floor plans. Generating seating scenarios to compare differing constraints necessitates the need for fast floor plan iteration. 

With this in mind, we limit the execution time of each \textsc{SA} sub-problem to a total of 10 minutes to mimic an online allocation scenario. Additionally, the \textsc{SA} sub-problem is solved with \textsc{ICA} plus Local Search (\textsc{ICS+LS}) and the number of neighbors is limited to $s_{n}=5$. 

Similarly, for the \textsc{IPSA} approach we do not report the standard deviation for the run time, as only one run was performed, due to it being solved to optimality. 
The complete results are reported in Table~\ref{tab:quantitative table}.

\begin{table*}[tbh]
    \small
    \centering

\begin{tabular}{@{}cccccccc@{}}
                                                            &   & \multicolumn{2}{c}{Small} & \multicolumn{2}{c}{Medium}  & \multicolumn{2}{c}{Large}  \\ \cmidrule(l){3-8} 
                                                                Method      &       Metrics     &       \multicolumn{6}{c}{Delayed Office Decision} \\ \cmidrule(l){3-8}     && True    &      False       &       True     &      False       &       True        & False                             \\ \midrule
\multirow{3}{2cm}{\centering \textbf{IPSA}}             & Central Seat Distance     &  318.8& 308.8          &  936.5   &    930.9   &  272.7    & 289.6\\
                                                            & Office Distance & 545.2 & 492.0  &  1076.4 & 1036.8 &  519.7&  479.5 \\
                                                            & Exec. Time (s)   &  $95.49	\!\pm\!-$           &  $139.0\!\pm\!-$ &  $629.5\!\pm\!-$  &  $647.8\!\pm\!-$ & $3748.4\!\pm\!-$ & $3708.5\!\pm\!-$ \\ \midrule
\multirow{3}{2cm}{\centering \textbf{Greedy Algorithm}}     & Central Seat Distance &  - & 438.8 & - &1506.9 & - & 	571.2 \\
                                                            & Office Distance & - & 882.2 & -  & 1664.1 & - & 963.8 \\
                                                            & Exec. Time (s)   &  - &  $1.2\!\pm\!0.3$ & - &  $1.5\!\pm\!0.4$& - &  $14.6\!\pm\!1.9$   \\ \midrule

\multirow{3}{2cm}{\centering \textbf{ICA}}   & Central Seat Distance &  337.9& 336.9  &  976.3& 	968.7 & 284.1& 286.2 \\
                                                                & Office Distance & 548.1& 518.4	  & 1075.3 & 1081.8 & 513.7& 604.5	 \\
                                                                & Exec. Time (s)   &  $6.2\!\pm\!0.5$ &  $2.8\!\pm\!0.6$ & $6.7\!\pm\!0.3$  & $2.6\!\pm\!0.5$ & $62.1\!\pm\!5.8$ &  $34.0\!\pm\!5.9$  \\ \midrule 
\multirow{3}{2cm}{\centering \textbf{ICA++}}   & Central Seat Distance & 331.9  & \textbf{329.6} & 969.25 & 965.2 & 	281.6 & 283.0	 \\
                                                                & Office Distance & 546.2 & 510.5 & 1091.4 & 1035.4 & 513.9 &	 618.6 \\
                                                                & Exec. Time (s)   &  $6.1\!\pm\!0.5$           &  $2.6\!\pm\!0.6$ & $6.5\!\pm\!0.3$ & $2.5\!\pm\!0.4$ & $60.7\!\pm\!5.5$ &$33.2\!\pm\!6.9$   \\ \midrule 
\multirow{2}{2cm}{\centering \textbf{ICA + LS}}   & Central Seat Distance &  \textbf{329.05} & 331.54 & \textbf{961.95} & \textbf{950.22} & \textbf{273.4} & \textbf{274.2}	 \\
                                                                & Office Distance & 551.3 & 506.9 & 1112.9&  1037.35  &  515.0&	540.8 \\
                                                                & Exec. Time (s)   &  $9.9\!\pm\!2.0$ &  $6.5\!\pm\!1.7$ & $8.3\!\pm\!0.4$ & $4.2\!\pm\!0.3$ & $956.2\!\pm\!207.9$ & $987.9\!\pm\!227.6$   \\ \bottomrule
                        
\end{tabular}
\caption{Quantitative evaluation of different approaches by using the \textsc{DF-HSA} algorithm. We highlight the lowest heuristic result in each column with bold. We report both the average execution time along with the standard error.}
\label{tab:quantitative table}
\end{table*}

\begin{figure*}[]
    \centering
    \begin{subfigure}[t]{0.31\textwidth}
        \centering
        \includegraphics[width=\textwidth]{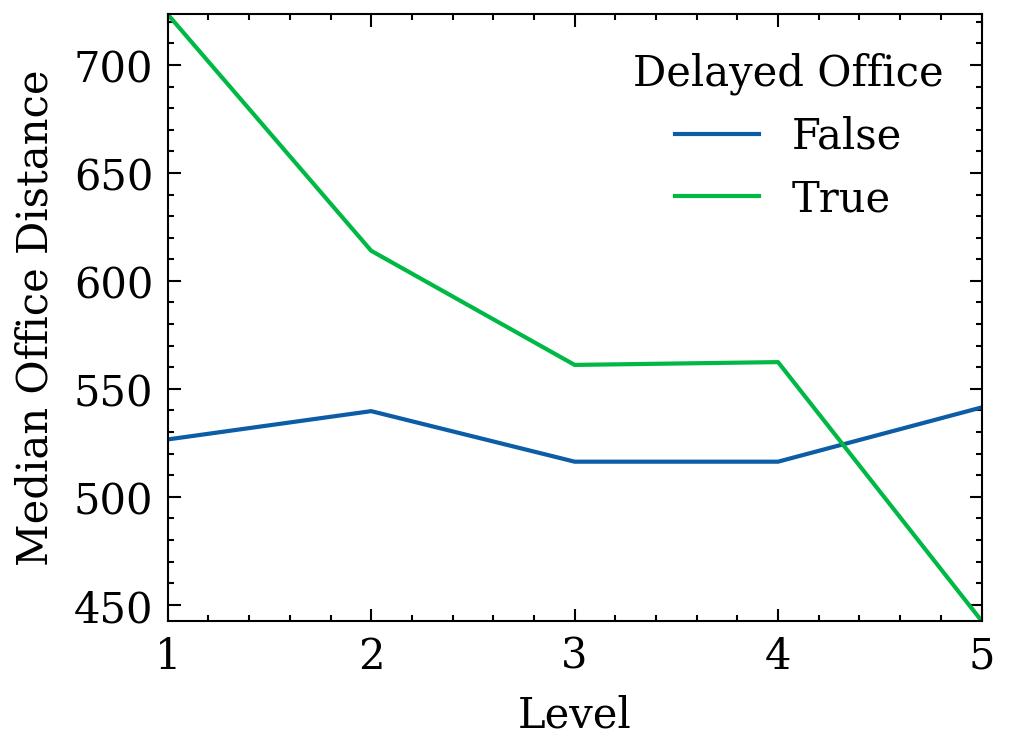}
        \caption{Small}
        \label{fig:383 delayed}
    \end{subfigure}
    \begin{subfigure}[t]{0.31\textwidth}
        \centering
        \includegraphics[width=\textwidth]{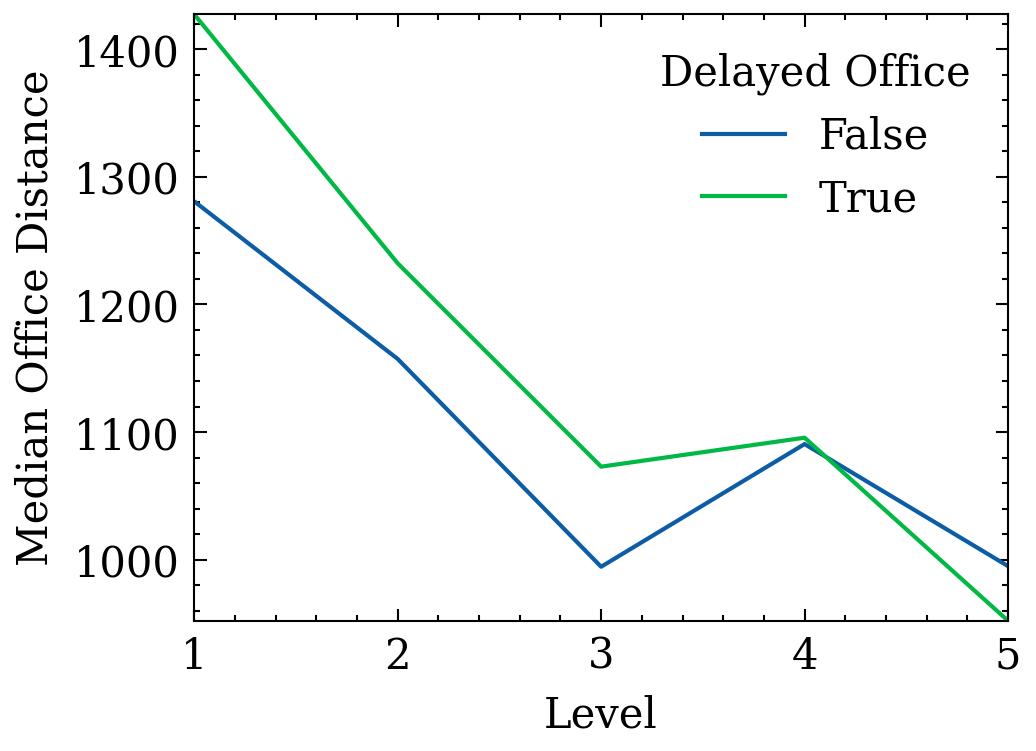}
        \caption{Medium}
        \label{fig:270 Park delayed}
    \end{subfigure}
    \begin{subfigure}[t]{0.31\textwidth}
        \centering
        \includegraphics[width=\textwidth]{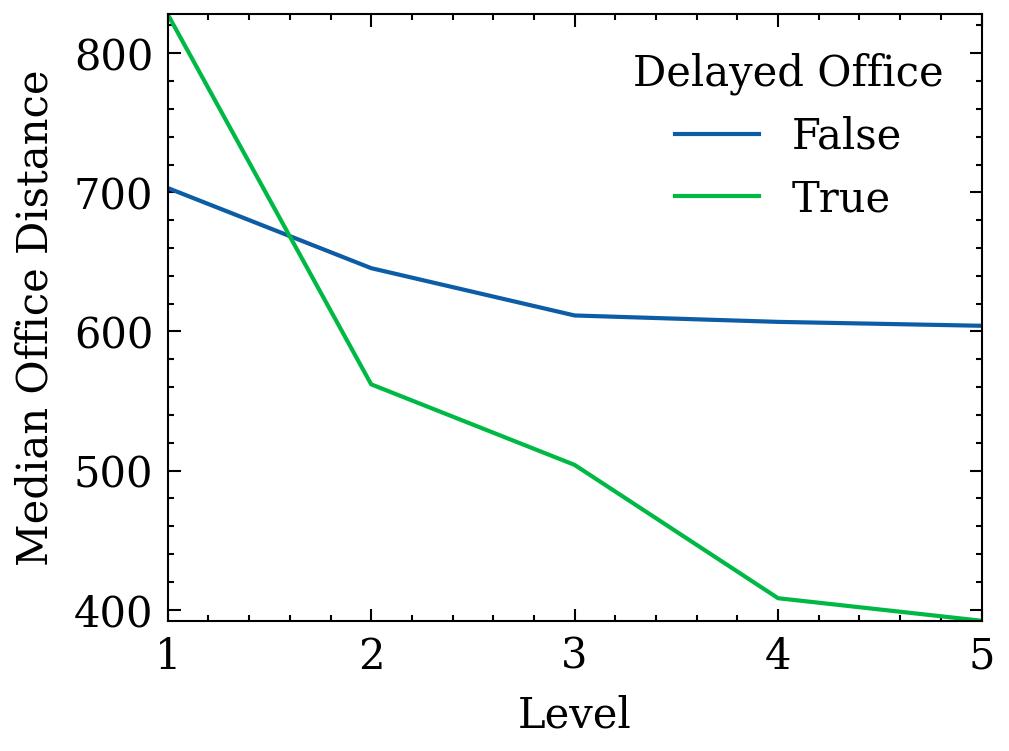}
        \caption{Large}
        \label{fig:25bs delayed}
    \end{subfigure}
    \caption{Depicts the effect on total office distance per level of \textsc{DF-HSA} across all methods. Note how delaying the office decision enables higher level hierarchy teams to on average sit closer to their aligned offices. This is beneficial as the purpose of the \textsc{HSA} framework is to allocate leaf teams while considering their branch parent's team.}
    \label{fig:combined}
\end{figure*}
The results shows the \textsc{IPSA} approach across all heuristic methods has the lowest total average distance to central seat across all levels and methods except for the largest instance.
The \textsc{IPSA} method performs worse in larger scale instances as the time limit is often reached before the optimality gap is closed by the solver due to its quadratic variable scaling in $|S|$.

Our results indicate that \textsc{ICA + LS} outperforms other heuristics, but at a slightly increased execution time cost due to the local search procedure in small and medium instance.
However, \textsc{ICA + LS} converges slowly in large instances as the local search IP procedure is based on the \textsc{IPSA} problem formulation which in our instances did not scale with increased sizing of $|S|$
Lastly, we observe that using the k-means++ approach slightly improves the objective function compared to the \textsc{ICA} approach with little to no cost to the execution time.
Empirically, varying $|S|$ or $|T|$ across different instances is insufficient to gauge the time-to-solution, as the difficulty and structure of the organizational hierarchy can vary unpredictably. Therefore, we cannot quantitatively conclude which method performs better across different cardinalities of $S$ or $T$.

\subsection{Qualitative Evaluation}\label{sec:eval:qual}

As stated earlier the human-evaluated solution quality is both affected by the estimation of pairwise seat distances and the team-to-seat allocation optimization. Therefore, we will conduct a qualitative evaluation of both contributions.
\begin{figure*}[ht]
    \centering
    \begin{subfigure}[t]{0.45\textwidth}
        \centering
        \includegraphics[width=\textwidth]{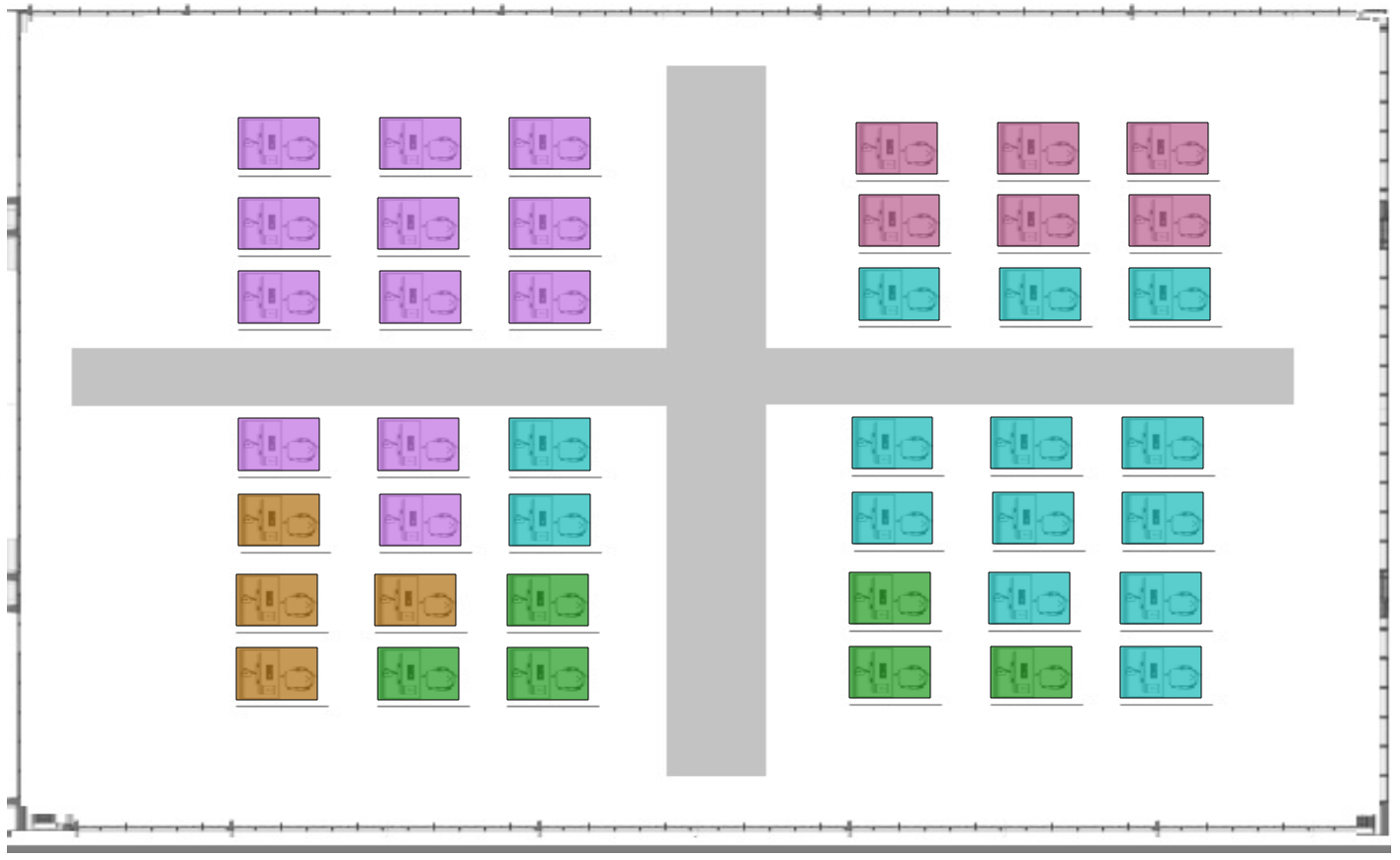}
        \caption{Euclidean Distance}
        
    \end{subfigure}
   \begin{subfigure}[t]{0.45\textwidth}
        \centering
        \includegraphics[width=\textwidth]{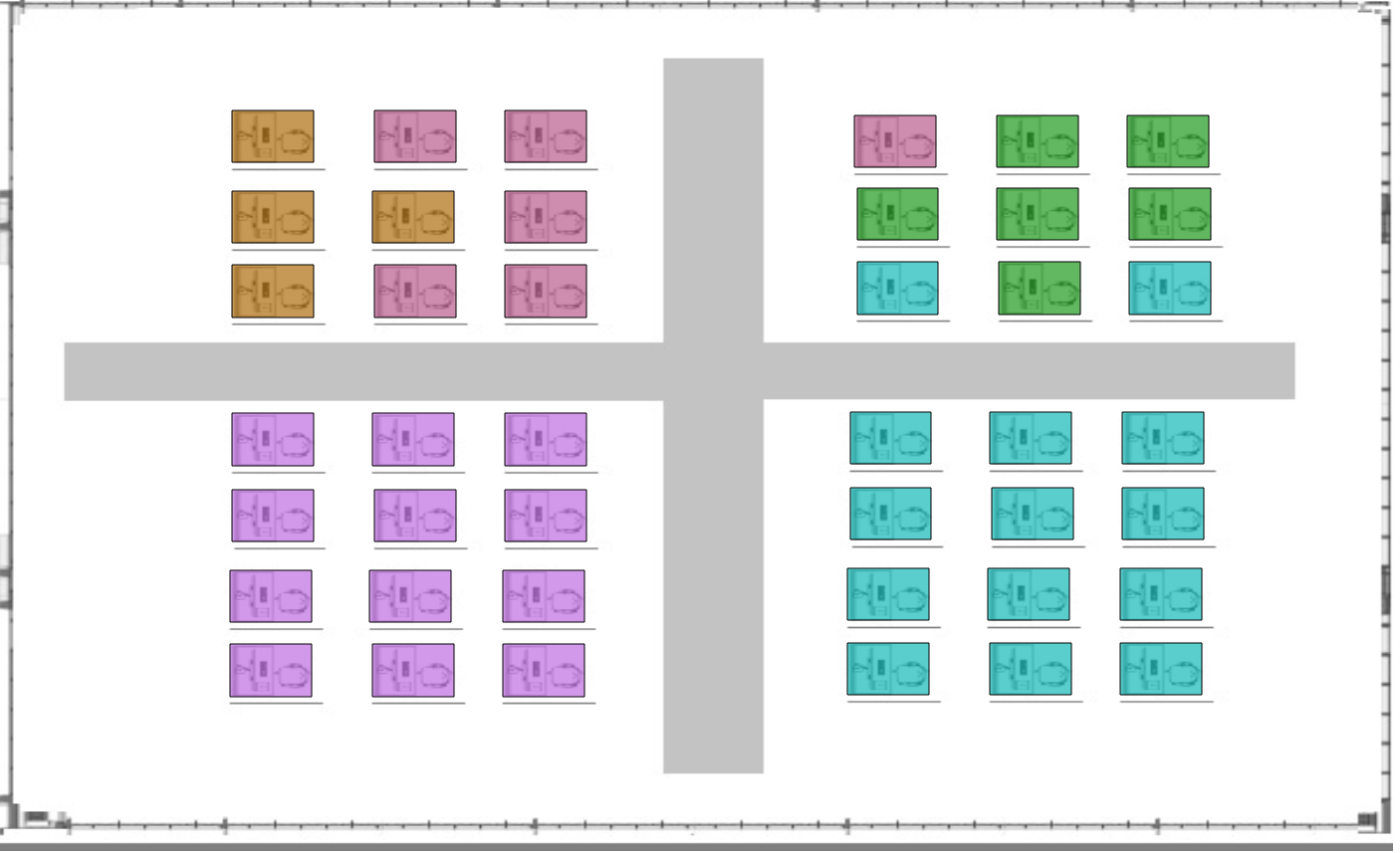}
        \caption{Allocation with PRM+RRT Distance}
        
    \end{subfigure}
    \caption{Toy example allocation of 4 teams being allocated to a total of 42 seats which are physically separated by four gray-colored walls using \emph{IPSA} framework with different seat distance estimation methods.}
    \label{fig:output_distance}
\end{figure*}

To evaluate the impact of an improved distance estimation framework for estimating pair-wise seat distances, we conducted a toy example as seen in Figure~\ref{fig:output_distance}. The setup of the toy example is that we have 4 teams and 4 separate areas divided by impassable walls. Two of the teams are too large for any given area and have to be split to one or more other areas by the allocation algorithm. Although the allocation solutions appear similar at first, the differences are significant. When using the Euclidean distance, more teams are split between areas due to the underestimation of seat distances caused by physically impassable walls. This effect is particularly evident in the green team, which is divided into a trapezoid shape in Figure~\ref{fig:output_distance}(a). Consequently, while the two optimization metrics yield similar numerical results the quality of the solution, as assessed by human evaluators, differs.

Additionally, Figure~\ref{fig:output_distance}(b) illustrates the gap between constraint optimization and human preferences. Although the optimization procedure was solved to optimality, a human decision-maker might choose to manually relocate the two teal seats in the top right quadrant to be adjacent, even if this adjustment slightly reduces the overall optimization objective.

Next, we will highlight a few of our observations related to the solution quality generated by the framework which serves as guidance to understanding the strengths and weaknesses of the methods proposed.
An example output of the \textsc{HSA} problem can be seen in Figure~\ref{fig:output}, where we have solved the \textsc{SA} sub-problems with \textsc{ICA+LS}. The figure shows the allocation of an organizational hierarchy with multiple teams to a floor plan. 

When qualitatively comparing the three \textsc{ICA} heuristics, we did not observe a significant difference in human-evaluated solution quality for small and medium-scale instances.

Interestingly, when pairwise seat distances are estimated using the \textsc{PRM} method, the resulting allocation forms a ball shape. This is because nodes are connected if they are within a $\delta_s$ radius of each other. Importantly, the framework used for distance estimation directly impacts the resulting allocation.
The ball-shaped output from using \textsc{PRM} may not be ideal for open and densely packed office floor plans. However, in floor plans with greater distances between seats, the ball-shaped allocation was less pronounced. In scenarios where seats are densely arranged in long rows, it may be more effective to explore approaches that aim to find the smallest bounding box for each team or another framework.

Future work should try to improve the disconnect between the constrained programming and human-evaluated solution quality by studying in which scenarios different solution methods and distance estimation techniques performs best.

\section{Conclusion}\label{sec:conclusion}

Our contributions are twofold: (1) we introduce a novel approach for calculating pairwise seat distances to overcome representation problems for high resolution floor plans, and (2) introduce the hierarchical seating allocation problem (\textsc{HSAP}). We compute pairwise seat distances by accounting for impassable physical walls using \textsc{PRM} and \textsc{RRT} methods. Additionally, we propose new techniques for considering organizational hierarchies to allocate seats to teams at any hierarchical level within a floor plan. The solution to the \textsc{HSAP} is formulated as a mixed-integer programming problem, which is decomposed into several sub-problems that can be efficiently solved as individual seating allocation (\textsc{SA}) problems.

We also introduce several innovative methods for solving a multi-resource \textsc{SA} problem, along with a local search heuristic that can significantly enhance incumbent solutions for small and medium-scale instances without substantially increasing execution time.

Future work could focus on developing a dynamic pipeline to determine the most suitable algorithm(s) based on the size and complexity of each subproblem could be beneficial. For the former, inspiration could be drawn from Gn\"agi and Baumann~\shortcite{Gngi2021} to create a more robust local search procedure. For the latter, the pipeline devised by Stefanello, de Araujo, and M\"uller~\shortcite{Stefanello2014} could be adapted for this purpose.

\section{\textbf{Disclaimer.}}
This paper was prepared for informational purposes by
the Artificial Intelligence Research group of JPMorgan Chase \& Co. and its affiliates (``JP Morgan''),
and is not a product of the Research Department of JP Morgan.
JP Morgan makes no representation and warranty whatsoever and disclaims all liability,
for the completeness, accuracy or reliability of the information contained herein.
This document is not intended as investment research or investment advice, or a recommendation,
offer or solicitation for the purchase or sale of any security, financial instrument, financial product or service,
or to be used in any way for evaluating the merits of participating in any transaction,
and shall not constitute a solicitation under any jurisdiction or to any person,
if such solicitation under such jurisdiction or to such person would be unlawful.
\\
\\
\textcopyright 2026 JPMorgan Chase \& Co. All rights reserved.
\clearpage
\bibliography{main}

@article{kav96_prm,
  author={Kavraki, L.E. and Svestka, P. and Latombe, J.-C. and Overmars, M.H.},
  journal={IEEE Transactions on Robotics and Automation}, 
  title={Probabilistic roadmaps for path planning in high-dimensional configuration spaces}, 
  year={1996},
  volume={12},
  number={4},
  pages={566-580}
}

@techreport{lav98_rrt,
  author={S. M. LaValle},
  title={Rapidly-exploring random trees: A new tool for path planning}, 
  institution={Iowa State University},
  year={1998}
}

@article{hechenberger2022multirayScan,
  title={Multi-target search in euclidean space with ray shooting (full version)},
  author={Hechenberger, Ryan and Harabor, Daniel and Cheema, Muhammad Aamir and Stuckey, Peter J and Bodic, Pierre Le},
  journal={arXiv preprint arXiv:2207.02436},
  year={2022}
}

@misc{cohen2017fastmapalgorithmshortestpath,
      title={The FastMap Algorithm for Shortest Path Computations}, 
      author={Liron Cohen and Tansel Uras and Shiva Jahangiri and Aliyah Arunasalam and Sven Koenig and T. K. Satish Kumar},
      year={2017},
      eprint={1706.02792},
      archivePrefix={arXiv},
      primaryClass={cs.AI},
      url={https://arxiv.org/abs/1706.02792}, 
}

@inproceedings{polyanya,
author = {Cui, Michael L. and Harabor, Daniel D. and Grastien, Alban},
title = {Compromise-free pathfinding on a navigation mesh},
year = {2017},
isbn = {9780999241103},
publisher = {AAAI Press},
booktitle = {Proceedings of the 26th International Joint Conference on Artificial Intelligence},
pages = {496–502},
numpages = {7},
location = {Melbourne, Australia},
series = {IJCAI'17}
}

@article{Hales2019,
  doi = {10.1007/s11750-019-00515-3},
  url = {https://doi.org/10.1007/s11750-019-00515-3},
  year = {2019},
  month = apr,
  publisher = {Springer Science and Business Media {LLC}},
  volume = {27},
  number = {3},
  pages = {426--455},
  author = {Roland Oliver Hales and Sergio Garc{\'{\i}}a},
  title = {Congress seat allocation using mathematical optimization},
  journal = {{TOP}}
}

@article{Wu2021,
  title = {Fast-RRT: A RRT-Based Optimal Path Finding Method},
  volume = {11},
  ISSN = {2076-3417},
  url = {http://dx.doi.org/10.3390/app112411777},
  DOI = {10.3390/app112411777},
  number = {24},
  journal = {Applied Sciences},
  publisher = {MDPI AG},
  author = {Wu,  Zhenping and Meng,  Zhijun and Zhao,  Wenlong and Wu,  Zhe},
  year = {2021},
  month = dec,
  pages = {11777}
}

@ARTICLE{A*,
  author={Hart, Peter E. and Nilsson, Nils J. and Raphael, Bertram},
  journal={IEEE Transactions on Systems Science and Cybernetics}, 
  title={A Formal Basis for the Heuristic Determination of Minimum Cost Paths}, 
  year={1968},
  volume={4},
  number={2},
  pages={100-107},
  doi={10.1109/TSSC.1968.300136}}

@article{Mu2019,
  title = {On solving large p-median problems},
  volume = {47},
  ISSN = {2399-8091},
  url = {http://dx.doi.org/10.1177/2399808319892598},
  DOI = {10.1177/2399808319892598},
  number = {6},
  journal = {Environment and Planning B: Urban Analytics and City Science},
  publisher = {SAGE Publications},
  author = {Mu,  Wangshu and Tong,  Daoqin},
  year = {2019},
  month = dec,
  pages = {981–996}
}

@article{Serafini2012,
  doi = {10.1016/j.mathsocsci.2011.08.006},
  url = {https://doi.org/10.1016/j.mathsocsci.2011.08.006},
  year = {2012},
  month = mar,
  publisher = {Elsevier {BV}},
  volume = {63},
  number = {2},
  pages = {107--113},
  author = {Paolo Serafini},
  title = {Allocation of the {EU} Parliament seats via integer linear programming and revised quotas},
  journal = {Mathematical Social Sciences}
}

@article{barry2021optimal,
  title={Optimal Seat Allocation Under Social Distancing Constraints},
  author={Barry, Michael and Gambella, Claudio and Lorenzi, Fabio and Sheehan, John and Ploennigs, Joern},
  journal={arXiv preprint arXiv:2105.05017},
  year={2021}
}

@inproceedings{kmeans++2007,
author = {Arthur, David and Vassilvitskii, Sergei},
title = {K-Means++: The Advantages of Careful Seeding},
year = {2007},
isbn = {9780898716245},
publisher = {Society for Industrial and Applied Mathematics},
address = {USA},
booktitle = {Proceedings of the Eighteenth Annual ACM-SIAM Symposium on Discrete Algorithms},
pages = {1027–1035},
numpages = {9},
location = {New Orleans, Louisiana},
series = {SODA '07}
}

@article{Gngi2021,
  doi = {10.1016/j.cor.2021.105304},
  url = {https://doi.org/10.1016/j.cor.2021.105304},
  year = {2021},
  month = aug,
  publisher = {Elsevier {BV}},
  volume = {132},
  pages = {105304},
  author = {Mario Gn\"{a}gi and Philipp Baumann},
  title = {A matheuristic for large-scale capacitated clustering},
  journal = {Computers and Operations Research}
}

@article{Mulvey1984,
  doi = {10.1016/0377-2217(84)90155-3},
  url = {https://doi.org/10.1016/0377-2217(84)90155-3},
  year = {1984},
  month = dec,
  publisher = {Elsevier {BV}},
  volume = {18},
  number = {3},
  pages = {339--348},
  author = {John M. Mulvey and Michael P. Beck},
  title = {Solving capacitated clustering problems},
  journal = {European Journal of Operational Research}
}

@article{Stefanello2014,
  doi = {10.1111/itor.12103},
  url = {https://doi.org/10.1111/itor.12103},
  year = {2014},
  month = jun,
  publisher = {Wiley},
  volume = {22},
  number = {1},
  pages = {149--167},
  author = {Fernando Stefanello and Olinto C. B. de Ara{\'{u}}jo and Felipe M. M\"{u}ller},
  title = {Matheuristics for the capacitated p-median problem},
  journal = {International Transactions in Operational Research}
}

@article{Stoll2022,

  url = {https://studenttheses.uu.nl/handle/20.500.12932/589},
    year={2022},
  author = {Stoll, Maximilian},
  title = {Solutions to the Distance Constrained
Cinema Seating Problem},
  journal = {Master Thesis, Utrecht University}
}

@article{Lorena2003,
  doi = {10.1023/a:1027353520175},
  url = {https://doi.org/10.1023/a:1027353520175},
  year = {2003},
  publisher = {Springer Science and Business Media {LLC}},
  volume = {3},
  number = {4},
  pages = {407--419},
  author = {Luiz Antonio Nogueira Lorena and Edson Luiz Fran{\c{c}}a Senne},
  title= {Local Search Heuristics for Capacitated p-Median Problems},
  journal = {Networks and Spatial Economics}
}

@article{Lorena2004,
  doi = {10.1016/s0305-0548(03)00039-x},
  url = {https://doi.org/10.1016/s0305-0548(03)00039-x},
  year = {2004},
  month = may,
  publisher = {Elsevier {BV}},
  volume = {31},
  number = {6},
  pages = {863--876},
  author = {Luiz A.N. Lorena and Edson L.F. Senne},
  title = {A column generation approach to capacitated p-median problems},
  journal = {Computers and Operations Research}
}

@article{Boccia2007,
  doi = {10.1007/s10852-007-9074-5},
  url = {https://doi.org/10.1007/s10852-007-9074-5},
  year = {2007},
  month = dec,
  publisher = {Springer Science and Business Media {LLC}},
  volume = {7},
  number = {1},
  pages = {43--58},
  author = {Maurizio Boccia and Antonio Sforza and Claudio Sterle and Igor Vasilyev},
  title = {A Cut and Branch Approach for the Capacitated p-Median Problem Based on Fenchel Cutting Planes},
  journal = {Journal of Mathematical Modelling and Algorithms}
}

@article{Reese2006,
  doi = {10.1002/net.20128},
  url = {https://doi.org/10.1002/net.20128},
  year = {2006},
  publisher = {Wiley},
  volume = {48},
  number = {3},
  pages = {125--142},
  author = {J. Reese},
  title = {Solution methods for thep-median problem: An annotated bibliography},
  journal = {Networks}
}

@article{Ceselli2005,
  doi = {10.1002/net.20059},
  url = {https://doi.org/10.1002/net.20059},
  year = {2005},
  publisher = {Wiley},
  volume = {45},
  number = {3},
  pages = {125--142},
  author = {Alberto Ceselli and Giovanni Righini},
  title = {A branch-and-price algorithm for the capacitated p-median problem},
  journal = {Networks}
}

@phdthesis{ulker2013office,
  title={Office space allocation by using mathematical programming and meta-heuristics},
  author={Ulker, Ozgur},
  year={2013},
  school={University of Nottingham}
}

@article{cplex2017v12,
  title={V12. 7: User’s Manual for CPLEX},
  author={CPLEX, IBM ILOG},
  journal={International Business Machines Corporation},
  year={2017}
}

@inproceedings{awadallah2012office,
  title={Office-space-allocation problem using harmony search algorithm},
  author={Awadallah, Mohammed A and Khader, Ahamad Tajudin and Al-Betar, Mohammed Azmi and Woon, Phuah Chea},
  booktitle={International Conference on Neural Information Processing},
  pages={365--374},
  year={2012},
}
\newpage
\onecolumn
\section{Appendix}\label{sec:Appendix}{}

\subsection{Appendix 1}\label{sec:Appendix1}
We define the binary variable $x_{t,j}$ as 1 if team $t$ is allocated seat $j$ and 0 otherwise. Similarly to Section~\ref{sec:method:seats}, we define $o_t$ and $d_t$ as the office and desk requirement of team $t$ respectively, and $S$ as the set of all seats and $S_d$ and $S_o$ as the set of desk and office respectively. This yields the following mixed integer quadratic (MIQP) formulation as it is quadratic in the objective:
\begin{align}
    \min \sum_{t \in T, k\in S,j \in S} D(k,j) x_{tk} x_{tj} \label{eq:MIQPObjective} \\
    \text{s.t.} \sum_{t\in T} x_{tk}\le 1 \quad \forall k \in S \label{eq:MIQPSeat} \\
    \sum_{k_s\in S_d} x_{tk_s}= d_t  \quad \forall t \in T \label{eq:MIQPTeamDeskCapacity}  \\
    \sum_{k_o\in S_o} x_{tk_o}=o_t \quad \forall t \in T \label{eq:MIQPTeamOfficeCapacity}
\end{align}

The constraints can be explained as follows:
\begin{itemize}
    \item The objective in Eq.~\ref{eq:MIQPObjective} minimizes the total pairwise distance to each allocated seat for each team.
    \item Eq.~\ref{eq:MIQPSeat} describes that a seat can only be allocated to at most one team.
    \item Eq.~\ref{eq:MIQPTeamDeskCapacity} and equation~\ref{eq:MIQPTeamOfficeCapacity} describes that each team must get allocated the number of desks and office to satisfy its demand for each of the respective resources.
\end{itemize}

\end{document}